# Eligibility Propagation to Speed up Time Hopping for Reinforcement Learning

Petar S. Kormushev, Kohei Nomoto, Fangyan Dong, and Kaoru Hirota

*Abstract*— **A mechanism called Eligibility Propagation is proposed to speed up the Time Hopping technique used for faster Reinforcement Learning in simulations. Eligibility Propagation provides for Time Hopping similar abilities to what eligibility traces provide for conventional Reinforcement Learning. It propagates values from one state to all of its temporal predecessors using a state transitions graph. Experiments on a simulated biped crawling robot confirm that Eligibility Propagation accelerates the learning process more than 3 times.**

## I. Introduction

REINFORCEMENT learning (RL) algorithms [16] address the problem of learning to select optimal actions when limited feedback (usually in the form of a scalar reinforcement function) from the environment is available.

General RL algorithms like Q-learning [17], SARSA and TD($\lambda$) [15] have been proved to converge to the globally optimal solution (under certain assumptions) [1][17]. They are very flexible, because they do not require a model of the environment, and have been shown to be effective in solving a variety of RL tasks. This flexibility, however, comes at a certain cost: these RL algorithms require extremely long training to cope with large state space problems.

Many different approaches have been proposed for speeding up the RL process. One possible technique is to use function approximation [8], in order to reduce the effect of the "curse of dimensionality". Unfortunately, using function approximation creates instability problems when used with off-policy learning.

Significant speed-up can be achieved when a demonstration of the goal task is available [3], as in Apprenticeship Learning [7]. Although there is a risk of running dangerous exploration policies in the real world [10], there are successful implementations of apprenticeship learning for aerobatic helicopter flight [11].

Another possible technique for speeding up RL is to use some form of hierarchical decomposition of the problem [4]. A prominent example is the "MAXQ Value Function Decomposition" [2]. Hybrid methods using both apprenticeship learning and hierarchical decomposition have been successfully applied to quadruped locomotion [14][18]. Unfortunately, decomposition of the target task is not always possible, and sometimes it may impose additional burden on the users of the RL algorithm.

A state-of-the-art RL algorithm for efficient state space exploration is E3 [6]. It uses active exploration policy to visit states whose transition dynamics are still inaccurately modeled. Because of this, running E3 directly in the real world might lead to a dangerous exploration behavior.

Instead of executing RL algorithms in the real world, simulations are commonly used. This approach has two main advantages: speed and safety. Depending on its complexity, a simulation can run many times faster than a real-world experiment. Also, the time needed to set up and maintain a simulation experiment is far less compared to a real-world experiment. The second advantage, safety, is also very important, especially if the RL agent is a very expensive equipment (e.g. a fragile robot), or a dangerous one (e.g. a chemical plant). Whether the full potential of computer simulations has been utilized for RL, however, is an open question.

A new trend in RL suggests that this might not be the case. For example, two techniques have been proposed recently to better utilize the potential of computer simulations for RL: Time Manipulation [12] and Time Hopping [13]. They share the concept of using the simulation time as a tool for speeding up the learning process. The first technique, called Time Manipulation, suggests that doing backward time manipulations inside a simulation can significantly speed up the learning process and improve the state space exploration. Applied to failure-avoidance RL problems, such as the cart-pole balancing problem, Time Manipulation has been shown to increase the speed of convergence by 260% [12].

This paper focuses on the second technique, called Time Hopping, which can be applied successfully to continuous optimization problems. Unlike the Time Manipulation technique, which can only perform backward time manipulations, the Time Hopping technique can make arbitrary "hops" between states and traverse rapidly throughout the entire state space. It has been shown to accelerate the learning process more than 7 times on some problems [13]. Time Hopping possesses mechanisms to trigger time manipulation events, to make prediction about possible future rewards, and to select promising time hopping targets.

This paper proposes an additional mechanism called Eligibility Propagation to be added to the Time Hopping

Manuscript submitted March 31, 2009. This work was supported in part by the Japanese Ministry of Education, Culture, Sports, Science and Technology (MEXT).

P. S. Kormushev, F. Dong and K. Hirota are with the Department of Computational Intelligence and Systems Science, Tokyo Institute of Technology, Yokohama, 226-8502, Japan. (phone: +81-45-924-5686/5682; fax: +81-45-924-5676; e-mail: {petar, tou, hirota}@hrt.dis.titech.ac.jp).

K. Nomoto is with the Industrial Design Center, Mitsubishi Electric Corporation, Tokyo, Japan.
(e-mail: Nomoto.Kohei@dw.MitsubishiElectric.co.jp)

technique, in order to provide similar abilities to what eligibility traces provide for conventional RL. Eligibility traces are easy to implement for conventional RL methods with sequential time transitions, but in the case of Time Hopping, due to its non-sequential nature, a number of obstacles have to be overcome.

The following Section II makes a brief overview of the Time Hopping technique and its components. Section III explains why it is important (and not trivial) to implement some form of eligibility traces for Time Hopping and proposes the Eligibility Propagation mechanism to do this. Section IV presents the results from experimental evaluation of Eligibility Propagation on a benchmark continuous-optimization problem: a biped crawling robot.

## II. OVERVIEW OF TIME HOPPING

### A. Basics of Time Hopping

Time Hopping is an algorithmic technique which allows maintaining higher learning rate in a simulation environment by hopping to appropriately selected states [13]. For example, let us consider a formal definition of a RL problem, given by the Markov Decision Process (MDP) on Fig. 1. Each state transition has a probability associated with it. State 1 represents situations of the environment that are very common and learned quickly. The frequency with which state 1 is being visited is the highest of all. As the state number increases, the probability of being in the corresponding state becomes lower. State 4 represents the rarest situations and therefore the most unlikely to be well explored and learned.

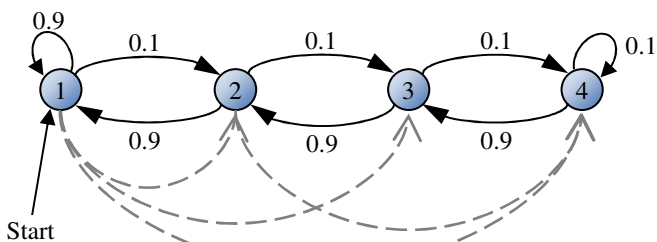

Fig. 1. An example of a MDP with uneven state probability distribution. Time Hopping can create "shortcuts in time" (shown with dashed lines) between otherwise distant states, i.e. states connected by a very low-probability path. This allows even the lowest-probability state 4 to be learned easily.

When applied to such a MDP, Time Hopping creates "shortcuts in time" by making hops (direct state transitions) between very distant states inside the MDP. Hopping to low-probability states makes them easier to be learned, while at the same time it helps to avoid unnecessary repetition of already well-explored states [13]. The process is completely transparent for the underlying RL algorithm.

### B. Components of Time Hopping

When applied to a conventional RL algorithm, Time Hopping consists of 3 components:

1) Hopping trigger – decides when the hopping starts;
2) Target selection – decides where to hop to;
3) Hopping – performs the actual hopping.

The flowchart on Fig. 2 shows how these 3 components of Time Hopping are connected and how they interact with the RL algorithm.

When the Time Hopping trigger is activated, a target state and time have to be selected, considering many relevant properties of the states, such as probability, visit frequency, level of exploration, connectivity to other states (number of state transitions), etc. After a target state and time have been selected, hopping can be performed. It includes setting the RL agent and the simulation environment to the proper state, while at the same time preserving all the acquired knowledge by the agent.

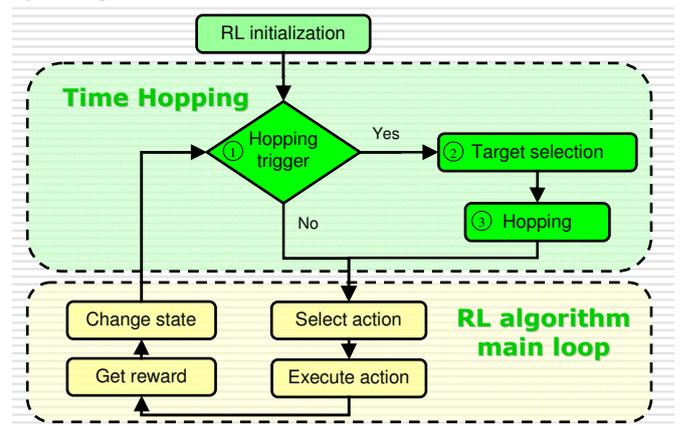

Fig. 2. Time Hopping technique applied to a conventional RL algorithm. The lower group (marked with a dashed line) contains the conventional RL algorithm main loop, into which the Time Hopping components (the upper group) are integrated.

## III. ELIGIBILITY PROPAGATION

### A. The role of eligibility traces

Eligibility traces are one of the basic mechanisms for temporal credit assignment in reinforcement learning [16]. An eligibility trace is a temporary record of the occurrence of an event, such as the visiting of a state or the taking of an action. When a learning update occurs, the eligibility trace is used to assign credit or blame for the received reward to the most appropriate states or actions. For example, in the popular TD($\lambda$) algorithm, the $\lambda$ refers to the use of an eligibility trace. Almost any temporal-difference (TD) methods, e.g., Q-learning and SARSA, can be combined with eligibility traces to obtain a more general method that may learn more efficiently. This is why it is important to implement some form of eligibility traces for Time Hopping as well, in order to speed up its convergence.

Eligibility traces are usually easy to implement for conventional RL methods. In the case of Time Hopping, however, due to its non-sequential nature, it is not trivial to do so. Since arbitrary hops between states are allowed, it is impossible to directly apply linear eligibility traces. Instead, we propose a different mechanism called Eligibility Propagation to do this.

### B. Eligibility Propagation mechanism

Time Hopping is guaranteed to converge when an off-policy RL algorithm is used [13], because the learned policy is independent of the policy followed during learning. This means that the exploration policy does not converge to the optimal policy. In fact, Time Hopping deliberately tries to avoid convergence of the policy in order to maintain high learning rate and minimize exploration redundancy. This poses a major requirement for any potential eligibility-trace-mechanism: it has to be able to learn from independent non-sequential state transitions spread sparsely throughout the state space.

The proposed solution is to construct an oriented graph which represents the state transitions with their associated actions and rewards and use this data structure to propagate the learning updates. Because of the way Time Hopping works, the graph might be disconnected, consisting of many separate connected components.

Regardless of the actual order in which Time Hopping visits the states, this oriented graph contains a record of the correct chronological sequence of state transitions. For example, each state transition can be considered to be from state $S_t$ to state $S_{t+1}$, and the information about this state transition is independent from what happened before it and what will happen after it. This allows to efficiently collect the separate pieces of information obtained during the randomized hopping, and to process them uniformly using the graph structure.

Once this oriented graph is available, it is used to propagate state value updates in the opposite direction of the state transition edges. This way, the propagation logically flows backwards in time, from state $S_t$ to all of its temporal predecessor states $S_{t-1}$, $S_{t-2}$ and so on. The propagation stops when the value updates become sufficiently small. The mechanism is illustrated on Fig. 3.

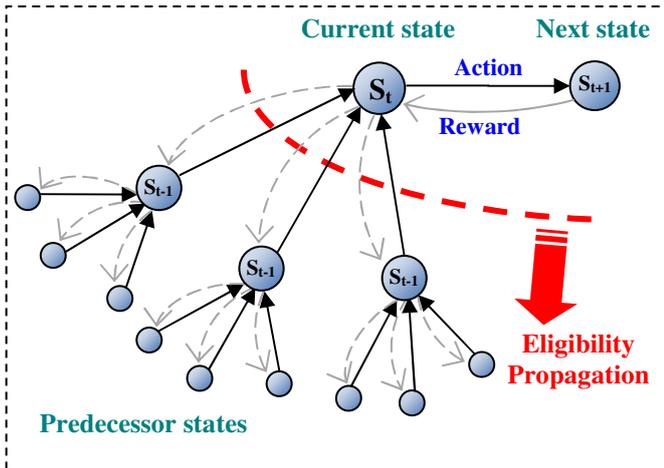

Fig. 3. Eligibility Propagation mechanism applied to the oriented graph of state transitions.

In summary, an explicit definition for the proposed mechanism is as follows:

Eligibility Propagation is an algorithmic mechanism for Time Hopping to efficiently collect, represent and propagate information about states and transitions. It uses a state transitions graph and a wave-like propagation algorithm to propagate state values from one state to all of its temporal predecessor states.

A concrete implementation of this mechanism within the Time Hopping technique is given in the following section.

### C. Implementation of Eligibility Propagation

The proposed implementation of Eligibility Propagation can be called "reverse graph propagation", because values are propagated inside the graph in reverse (opposite) direction of the state transitions' directions. The process is similar to the wave-like propagation of a BFS (breadth-first search) algorithm.

In order to give a more specific implementation description, Q-learning is used as the underlying RL algorithm. The following is the pseudo-code for the proposed Eligibility Propagation mechanism:

1. Construct an ordered set (queue) of state transitions called *PropagationQueue* and initialize it with the current state transition $\langle S_t, S_{t+1}\rangle$ in this way:

    $$PropagationQueue = \langle\langle S_t, S_{t+1}\rangle\rangle. \qquad (1)$$

2. Take the first state transition $\langle S_t, S_{t+1}\rangle \in PropagationQueue$ and remove it from the queue.

3. Let $Q_{\max}$ be the current maximum Q-value of state $S_t$:

    $$Q_{\max} = \max_A \{Q_{S_t, A}\}, \qquad (2)$$

    where the transition from state $S_t$ to state $S_{t+1}$ is done by executing action $A$, and the reward $R_{S_t, A}$ is received.

4. Update the Q-value for making the state transition $\langle S_t, S_{t+1}\rangle$ using the update rule:

    $$Q_{S_t, A} = R_{S_t, A} + \gamma \max_{A'}\{Q_{S_{t+1}, A'}\}. \qquad (3)$$

5. Let $Q'_{\max}$ be the new maximum Q-value of state $S_t$, calculated using formula (2).

6. If $|Q'_{\max} - Q_{\max}| > \varepsilon$, $\qquad (4)$

    then construct the set of all immediate predecessor state transitions of state $S_t$:

    $$\{\langle S_{t-1}, S_t\rangle \mid \langle S_{t-1}, S_t\rangle \in \text{transitions graph}\}, \qquad (5)$$

    and append it to the end of the *PropagationQueue*.

7. If $PropagationQueue \neq \langle\ \rangle$ then go to step 2.

8. Stop.

The decision whether further propagation is necessary is made in step 6. The propagation continues one more step backwards in time only if there is a significant difference between the old maximum Q-value and the new one, according to formula (4). This formula is based on the fact that $Q'_{\max}$ might be different than $Q_{\max}$ in exactly 3 out of 4 possible cases, which are:

- The transition $\langle S_t, S_{t+1}\rangle$ was the one with the highest value for state $S_t$ and its new (bigger) value needs to be propagated backwards to its predecessor states.

- The transition $\langle S_t, S_{t+1} \rangle$ was the one with the highest value but it is not any more, because its value is reduced. Propagation of the new maximum value (which belongs to a different transition) is necessary.
- The transition $\langle S_t, S_{t+1} \rangle$ was not the one with the highest value but now it became one, so its value needs propagation.

The only case when propagation is not necessary is when the transition $\langle S_t, S_{t+1} \rangle$ was not the one with the highest value and it is still not the one after the update. In this case, $Q'_{max}$ is equal to $Q_{max}$ and formula (4) correctly detects it and skips propagation.

In the previous 3 cases the propagation is performed, provided that there is a significant change of the value, determined by the $\varepsilon$ parameter. When $\varepsilon$ is smaller, the algorithm tends to propagate further the value changes. When $\varepsilon$ is bigger, it tends to propagate only the biggest changes just a few steps backwards, skipping any minor updates.

The depth of propagation also depends on the discount factor $\gamma$. The bigger $\gamma$ is, the deeper the propagation is, because longer-term reward accumulation is stimulated. Still, due to the exponential attenuation of future rewards, the $\gamma$ discount factor prevents the propagation from going too far and reduces the overall computational cost.

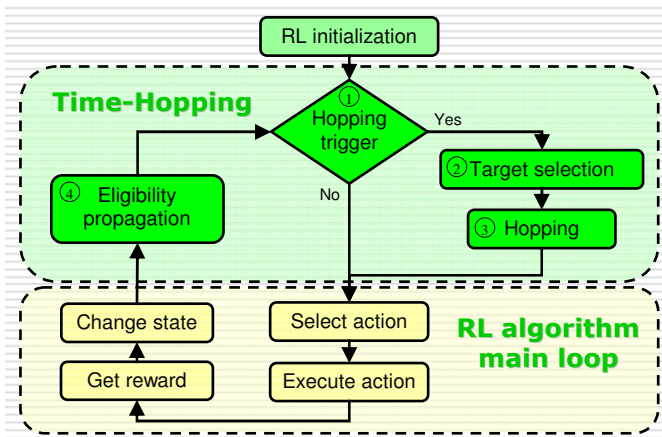

Fig. 4. Eligibility Propagation integrated as a 4th component in the Time-Hopping technique.

The described Eligibility Propagation mechanism can be encapsulated as a single component and integrated into the Time Hopping technique as shown on Fig. 4. It is called immediately after a state transition takes place, in order to propagate any potential $Q$-value changes, and before a time hopping step occurs.

## IV. APPLICATION OF ELIGIBILITY PROPAGATION TO BIPED CRAWLING ROBOT

In order to evaluate the efficiency of the proposed Eligibility Propagation mechanism, experiments on a simulated biped crawling robot are conducted. The goal of the learning process is to find a crawling motion with the maximum speed. The reward function for this task is defined as the horizontal displacement of the robot after every action.

### A. THEN experimental environment

A dedicated experimental software system called THEN (**T**ime **H**opping **EN**vironment) was developed for the purpose of this evaluation. A general view of the environment is shown on Fig. 5. THEN has a built-in physics simulation engine, implementation of the Time Hopping technique, useful visualization modules (for the simulation, the learning data and the state transitions graph) and most importantly – a prototype implementation of the Eligibility Propagation mechanism. To facilitate the analysis of the algorithm behavior, THEN displays detailed information about the current state, the previous state transitions, a visual view of the simulation, and allows runtime modification of all important parameters of the algorithms and the simulation. There is a manual and automatic control of the Time Hopping technique, as well as visualization of the accumulated data in the form of charts.

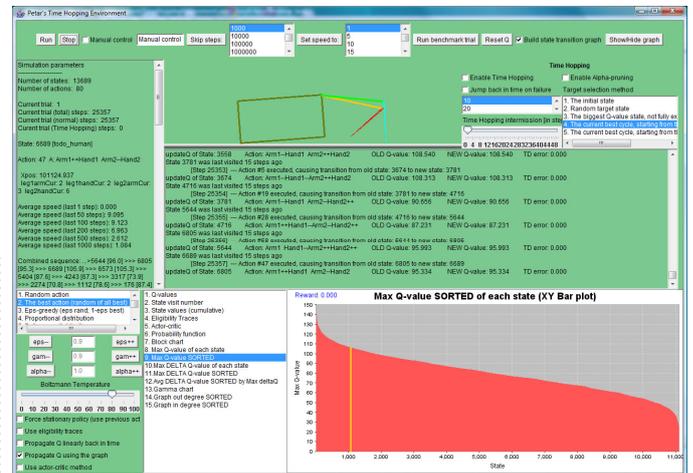

Fig. 5. General view of THEN (**T**ime **H**opping **EN**vironment). The built-in physics engine is running a biped crawling robot simulation.

### B. Description of the crawling robot

The experiments are conducted on a physical simulation of a biped crawling robot. The robot has 2 limbs, each with 2 segments, for a total of 4 degrees of freedom (DOF). Every DOF is independent from the rest and has 3 possible actions at each time step: to move clockwise, to move anti-clockwise, or to stand still. Fig. 6 shows a typical learned crawling sequence of the robot as visualized in the simulation environment constructed for this task.

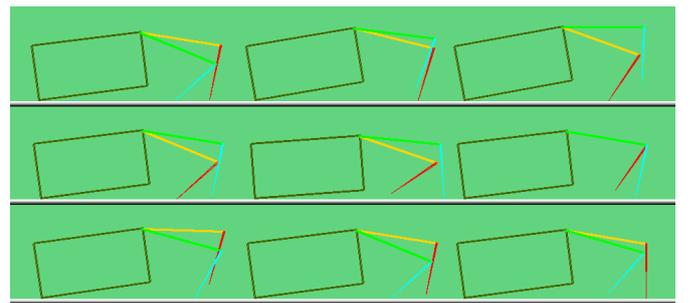

Fig. 6. Crawling robot with 2 limbs, each with 2 segments for a total of 4 DOF. Nine different states of the crawling robot are shown from a typical learned crawling sequence.

When all possible actions of each DOF of the robot are combined, assuming that they can all move at the same time independently, it produces an action space with size $3^4 - 1 = 80$ (we exclude the possibility that all DOF are standing still). Using appropriate discretization for the joint's angles (9 for the upper limbs and 13 for the lower limbs), the state space becomes divided into $(9 \times 13)^2 = 13689$ states. For better analysis of the crawling motion, each limb has been colored differently and only the "skeleton" of the robot is displayed.

### C. Description of the experimental method

The conducted experiments are divided in 3 groups: experiments using conventional Q-learning, experiments using only the Time Hopping technique applied to Q-learning (as described in [13]), and experiments using Time Hopping with Eligibility Propagation. The implementations used for the Time Hopping components are shown in Table I.

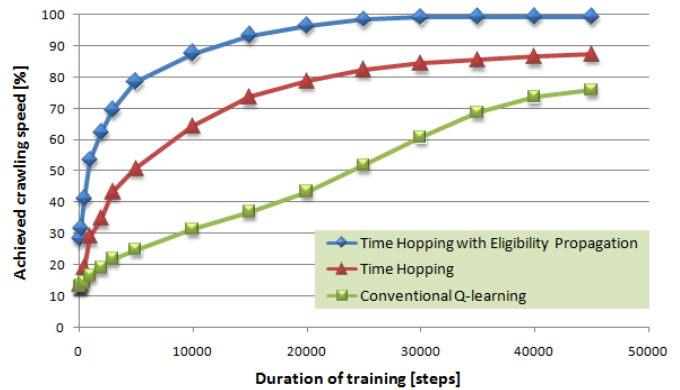

Fig. 7. Speed-of-learning comparison of conventional Q-learning, Time Hopping, and Time Hopping with Eligibility Propagation. It is based on the best solution achieved relative to the duration of training. The achieved crawling speed is measured as a percentage of the globally optimal solution, i.e. the fastest possible crawling speed of the robot.

TABLE I
IMPLEMENTATION USED FOR EACH TIME HOPPING COMPONENT

|   | Component name | Implementation used |
|---|---|---|
| 1 | Hopping trigger | Gamma pruning |
| 2 | Target selection | Lasso target selection |
| 3 | Hopping | Basic hopping |
| 4 | Eligibility propagation | Reverse graph propagation |

The experiments from all three groups are conducted in exactly the same way, using the same RL parameters (incl. discount factor γ, learning rate α, and the action selection method parameters). The initial state of the robot and the simulation environment parameters are also equal. The robot training continues up to a fixed number of steps (45000), and the achieved crawling speed is recorded at fixed checkpoints during the training. This process is repeated 10 times and the results are averaged, in order to ensure statistical significance.

### D. Evaluation of Eligibility Propagation

The evaluation of Eligibility Propagation is done using 3 main experiments.

In the first experiment, the learning speed of conventional Q-learning, Time Hopping, and Time Hopping with Eligibility Propagation is compared based on the best solution found (i.e. the fastest achieved crawling speed) for the same number of training steps. The comparison results are shown in Fig. 7. It shows the duration of training needed by each of the 3 algorithms to achieve a certain crawling speed. The achieved speed is displayed as percentage from the globally optimal solution.

The results show that Time Hopping with Eligibility Propagation is much faster than Time Hopping alone, which in turn is much faster than conventional Q-learning.

Compared to Time Hopping alone, Eligibility Propagation achieves significant speed-up of the learning process. For example, an 80%-optimal crawl is learned in only 5000 steps when Eligibility Propagation is used, while Time Hopping alone needs around 20000 steps to learn the same, i.e. in this case Eligibility Propagation needs 4 times fewer training steps to achieve the same result. The speed-up becomes even higher as the number of training steps increases. For example, Time Hopping with Eligibility Propagation reaches 90%-optimal solution with 12000 steps, while Time Hopping alone needs more than 50000 steps to do the same.

Compared to conventional Q-learning, Eligibility Propagation achieves even higher speed-up. For example, it needs only 4000 steps to achieve a 70%-optimal solution, while conventional Q-learning needs 36000 steps to learn the same, i.e. in this case Eligibility Propagation is 9 times faster. Time Hopping alone also outperforms conventional Q-learning by a factor of 3 in this case (12000 steps vs. 36000 steps).

In the second experiment, the real computational time of conventional Q-learning, Time Hopping, and Time Hopping with Eligibility Propagation is compared. The actual execution time necessary for each of the 3 algorithms to reach a certain crawling speed is measured.

The comparison results are shown in Fig. 8.

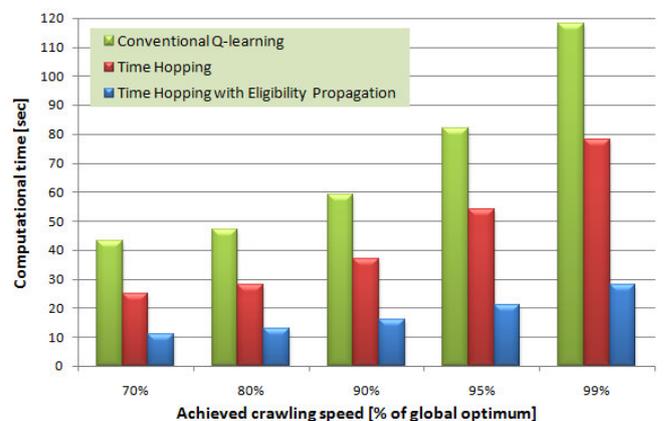

Fig. 8. Computational-time comparison of conventional Q-learning, Time Hopping, and Time Hopping with Eligibility Propagation. It is based on the real computational time of each algorithm required to reach a certain quality of the solution, i.e. certain crawling speed.

The results show that Time Hopping with Eligibility Propagation achieves 99% of the maximum possible speed almost 3 times faster than Time Hopping alone, and more than 4 times faster than conventional Q-learning. This significant speed-up of the learning process is achieved despite the additional computational overhead of maintaining the transitions graph. The reason for this is the improved Gamma-pruning based on more precise future reward predictions, as confirmed by the third experiment.

The goal of this third experiment is to provide insights about the state exploration and Q-value distribution, in order to explain the results from the previous two experiments. Conventional Q-learning, Time Hopping, and Time Hopping with Eligibility Propagation are compared based on the maximum Q-values achieved for all explored states after 45000 training steps. The Q-values are sorted in decreasing order and represent the distribution of Q-values within the explored state space. Fig. 9 shows the comparison results.

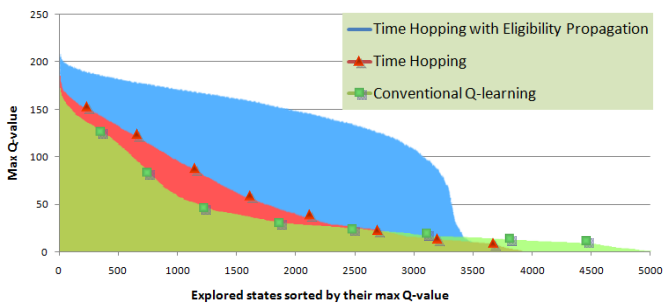

Fig. 9. State-exploration comparison of conventional Q-learning, Time Hopping, and Time Hopping with Eligibility Propagation. It shows the sorted sequence of maximum Q-values of all explored states after 45000 steps of training. Time Hopping with Eligibility Propagation has managed to find much higher maximum Q-values for the explored states. The conventional Q-learning has explored more states, but has found lower Q-values for them.

Firstly, the results show that Time Hopping with Eligibility Propagation has managed to find significantly higher maximum Q-values for the explored states compared to both conventional Q-learning and Time Hopping. The reason for this is that Eligibility Propagation manages to propagate well the state value updates among all explored states, therefore raising their maximum Q-values.

Secondly, the results show that both Time Hopping and Time Hopping with Eligibility Propagation have explored much fewer states than conventional Q-learning. The reason for this is the Gamma-pruning component of Time Hopping. It focuses the exploration of Time Hopping to the most promising branches and avoids unnecessary exploration. Conventional Q-learning does not have such a mechanism and therefore it explores more states, but finds lower Q-values for them.

Also, Time Hopping with Eligibility Propagation has explored slightly fewer states than Time Hopping alone. The reason for this is that while both algorithms concentrate the exploration on the most promising parts of the state space, only the Eligibility Propagation manages to propagate well the Q-values among the explored states. This improves the accuracy of the future reward estimation performed by the Gamma-pruning component of Time Hopping, which in its turn detects better unpromising branches of exploration and triggers a time hopping step to avoid them.

The more purposeful exploration and better propagation of the acquired state information help Eligibility Propagation to make the best of every single exploration step. This is a very important advantage of the proposed mechanism, especially if the simulation involved is computationally expensive. In this case, Eligibility Propagation can save real computational time by reducing the number of normal transition (simulation) steps in favor of Time Hopping steps.

## V. Conclusion

The Eligibility Propagation mechanism is proposed to provide for Time Hopping similar abilities to what eligibility traces provide for conventional RL.

During operation, Time Hopping completely changes the normal sequential state transitions into a rather randomized hopping behavior throughout the state space. This poses a challenge how to efficiently collect, represent and propagate knowledge about actions, rewards, states and transitions. Since using sequential eligibility traces is impossible, Eligibility Propagation uses the transitions graph to obtain all predecessor states of the updated state. This way, the propagation logically flows backwards in time, from one state to all of its temporal predecessor states.

The proposed mechanism is implemented as a fourth component of the Time Hopping technique. This maintains the clear separation between the 4 Time Hopping components and makes it straightforward to experiment with alternative component implementations.

The biggest advantage of Eligibility Propagation is that it can speed up the learning process of Time Hopping more than 3 times. This is due to the improved Gamma-pruning ability based on more precise future reward predictions. This, in turn, increases the exploration efficiency by better avoiding unpromising branches and selecting more appropriate hopping targets.

The conducted experiments on a biped crawling robot also show that the speed-up is achieved using significantly fewer training steps. As a result, the speed-up becomes even higher when the simulation is computationally more expensive, due to the more purposeful exploration. This property makes Eligibility Propagation very suitable for speeding up complex learning tasks which require costly simulation.

Another advantage of the proposed implementation of Eligibility Propagation is that no parameter tuning is necessary during the learning, which makes the mechanism easy to use.

Finally, an important drawback of the proposed technique is that it needs additional memory to store the transitions graph data. In other words, the speed-up is achieved by using more memory.